\documentclass[10pt,conference]{IEEEtran}
\IEEEoverridecommandlockouts
\usepackage{cite}
\usepackage{amsmath,amssymb,amsfonts}
\usepackage{algorithmic}
\usepackage{graphicx}
\usepackage{textcomp}
\usepackage{xcolor}
\usepackage{url}
\usepackage{xcolor}
\usepackage{soul}
\soulregister\cite7
\soulregister\ref7
\soulregister\item7
\soulregister\itemize7
\soulregister\pageref7
\soulregister\url7

\newcommand{\tensor}{\mathsf}

\newcommand{\netname}{\textit{pico}EmergencyNet }
\newcommand{\netnamenospace}{\textit{pico}EmergencyNet}

\def\BibTeX{{\rm B\kern-.05em{\sc i\kern-.025em b}\kern-.08em
    T\kern-.1667em\lower.7ex\hbox{E}\kern-.125emX}}

\begin{document}


\title{\netnamenospace: Attention-Enhanced Ultra-Small Deep Learning Model for Aerial Image Classification in Emergency Monitoring
}

 \author{\IEEEauthorblockN{Christos Kyrkou}
 \IEEEauthorblockA{\textit{KIOS Research and Innovation Center of Excellence} \\
 \textit{University of Cyprus}\\
 Nicosia, Cyprus \\
ORCID: 0000-0002-7926-7642}
 }

\maketitle

\begin{abstract}
Real-time disaster recognition from aerial imagery requires deep learning models that operate under strict memory, compute, and latency constraints. While recent approaches achieve high accuracy, their computational cost limits deployment on resource-constrained UAV platforms. This work introduces picoEmergencyNet, an ultra-lightweight convolutional neural network designed for aerial image disaster classification. The proposed architecture combines depthwise dilated convolutions with efficient channel attention and channel shuffling to enable multi-scale feature extraction at minimal computational cost. picoEmergencyNet has 23k parameters and 7M FLOPs, requiring only 0.1 MB of model storage. Evaluated on the AIDER and AIDERv2 disaster datasets, it achieves up to 96.1\% accuracy, matching or exceeding state-of-the-art lightweight models that are orders of magnitude larger. Deployment experiments on the NVIDIA Jetson AGX Orin show that, without hardware-specific optimizations, picoEmergencyNet achieves 258 FPS on GPU and 46 FPS on CPU with a 7× lower activation memory footprint compared to competing models. These results demonstrate that careful architectural specialization, rather than model scaling, enables highly efficient and accurate TinyML-ready disaster recognition for real-time UAV-based emergency monitoring.
\end{abstract}


\section{Introduction}
Advances in technologies such as unmanned aerial vehicles (UAVs) \cite{9979171}, have expanded the capabilities of disaster monitoring and response. UAVs provide quick aerial observations in hard-to-reach areas, which may be dangerous for humans to access, and capture high-resolution images, thus enhancing situational awareness over large areas. At the same time, Computer Vision (CV) algorithms based on deep learning techniques can serve as a crucial component for classifying disaster events by analyzing acquired images in real time, either during or in the aftermath of disasters. Consequently, the integration of UAVs with deep learning techniques for disaster recognition holds great promise for further advancements in the field of emergency management \cite{9979171}.

In recent times, several efforts have been made towards innovative approaches, which led to the establishment of multiple datasets designed for disaster classification \cite{kyrkou2020emergencynet,shianios2024direcnetv2}. Using these datasets multiple networks have been proposed based on Convolutional Neural Networks (CNNs or ConvNets) and lately Vision Transformers (ViTs). ConvNets and ViTs usually differ in their size and complexity, with standard ViTs often being more resource-intensive than ConvNets. 

Significantly, it should be highlighted that while deeper and more complex models tend to be more accurate, they come at additional cost of increased computational requirements, longer inference times, and heavier workloads. Such constraints become particularly critical in scenarios like emergency and disaster response. In scenarios where drone technology is frequently utilized, deploying such deep learning models for disaster response requires them to be compact and built with low-resource constraints in mind. This ensures reduced computational complexity, facilitating rapid inference and allowing real-time image analysis \cite{bouguettaya2019survey}. 

This work proposes an ultra-small deep learning model suitable for aerial image classification of disasters in emergency monitoring. Specifically, our main contributions are the following: 

\begin{itemize}
\item A tiny deep learning model that features minimal branching of depthwise dilated convolutions leading to limited activation memory and improved batch-1 latency, without exploiting hardware-specific optimizations.
\item An attention-based design to better mix dilated convolutions of multiple branches to allow cross-scale interaction leading to improved accuracy. 
\end{itemize}

These architectural considerations result in a proposed model referred to as \netname that achieves state-of-the-art accuracy while being compact (23k parameters, 7M FLOPs) and consistently outperforming other models on both GPU and CPU during real-world deployment.

\section{Related Works}

UAVs have become an important sensing platform for emergency management because they can be deployed rapidly and capture detailed views of affected areas. Recent work shows that deep-learning–based computer vision can support several UAV disaster applications, including real-time search and rescue operations \cite{9979171}, post-event damage recognition and damage-level classification from UAV imagery \cite{kyrkou2020emergencynet}. In these types of applications the machine learning and computer vision algorithms underpinning these systems not only need to provide sufficient accuracy but also be able to operate under tight time constraints, which makes efficiency and onboard performance central design goals. Therefore, recent surveys highlight the strong focus on lightweight models, optimization, and practical deployment constraints for real-time aerial vision \cite{habash2025aerial_od_survey}. 

Recent methods for disaster recognition systems heavily utilize deep learning methods such as Deep Convolutional Neural Networks \cite{krizhevsky2017imagenet} and Vision Transformers \cite{dosovitskiy2020image}. Since higher accuracy often comes with increased latency and resource use, disaster-response UAV deployments require lightweight, efficient models to enable fast, real-time inference \cite{bouguettaya2019survey}. 

Lightweight ConvNet models have become a focal point due to their compatibility with embedded systems, such as UAVs, which typically have limited computational capabilities. Initial methods utilized and repurposed existing models such as \cite{sandler2018mobilenetv2} that popularized depthwise separable convolutions, \cite{howard2019searching} that utilized automated network search, \cite{ma2018shufflenet} that introduced operations more suited for CPU platforms, and \cite{iandola2016squeezenet} that utilized multiple branches and channel reduction for efficient operation.

More specialized models derived from bottom-up have also been developed for the task of disaster recognition. Kyrkou et al. introduced EmergencyNet \cite{kyrkou2020emergencynet}, a small convolutional model based on depthwise convolutions, that demonstrated the applicability of specialized models for aerial disaster recognition. This work was then used as basis to develop TinyEmergencyNet \cite{mogaka2024tinyemergencynet} from the initial EmergencyNet  model through a pruning strategy that removed up to 60\% of the model weights of. \cite{lee2023watt} presented WATT-EffNet, an innovative approach improving accuracy while maintaining a lightweight architecture, with an F1-score of 0.885 on the AIDER dataset \cite{9025436}. \cite{munsif2022lightweight} proposed a 3 MB small model with an F1 score of 0.88 for automatic disaster recognition. \cite{yang2022efficient} optimized their model using OpenVINO, outperforming TensorFlow in frame-rate inference with minimal drop in accuracy. DiRecNetV2 proposed in \cite{shianios2024direcnetv2}, is a hybrid CNN–Transformer model for aerial disaster recognition from UAV imagery. The key idea is to combine CNNs (for efficient local feature extraction) with Transformer attention (for global context modeling) while keeping computational cost low enough for real-time deployment on drones. DiRecNetV2 achieves high single-label classification performance (weighted F1=0.964) and reasonable multi-label performance (F1=0.614). The work highlights the trade-off between accuracy, contextual modeling, and real-time constraints in UAV-based disaster response. TakuNet \cite{rossi2025takunet} proposes an ultra-lightweight CNN architecture for real-time disaster recognition on embedded UAV platforms. The model uses a mixture of efficiency-oriented design choices (depth-wise convolutions) along with platform-specific optimizations (TensorRT and FP16 inference) to minimize computation and energy consumption while maintaining competitive accuracy. Evaluated on aerial emergency datasets and deployed on embedded devices TakuNet achieves high accuracy with extremely high throughput demonstrating the benefits of small specialized models.

Within the domain of specific disaster recognition,  \cite{khudayberdiev2022fire} effectively utilized ViTs for fire detection, while \cite{ghali2021wildfire} implemented ViTs for wildfire segmentation. The research community has also explored ways to construct lightweight versions of ViTs. The goal of these efforts is to reduce the computing footprint of ViTs while maintaining their advantages, such as their adaptability and global receptive field. Thus, the following lightweight ViTs are additionally investigated; MobileViT-(s,xs,xss) \cite{mehta2021mobilevit}, and MobileViTV2-(050,100) \cite{mehta2206separable}.

This study aims to address the accuracy gap observed by small and tiny models and their larger counterparts while maintaining high performance.

\section{\netname}

This section describes \netnamenospace. A small neural network that applies attention across mixed dilated convolution channels to better blend multi-scale information. Furthermore, by making use of depthwise convolutions it leads to reduced parameter count and higher computational efficiency. 

The core block of the proposed network is motivated and built upon on ideas from multi-scale context possible via dilated convolutions \cite{chen2017deeplab}, the multi-path architecture of inception in \cite{szegedy2015going}, and depthwise convolutions \cite{sandler2018mobilenetv2}. The combination of these features and the incorporation of attention results in a block that is tailored for low-resource settings yet highly performant. Specifically, the block extracts multi-scale features that jointly capture fine-grained local details and broader global context, mixing and aggregating features and increasing their dimensionality via concatenation without incurring additional computational cost. Furthermore it leverages dilated convolutions to expand the receptive field by skipping input values, thereby covering a larger spatial area with fewer operations while yielding a richer representation. Depth-wise convolutions are used throughout all dilated stages to maintain computational efficiency. 

\subsection{Multi-Scale Context Extraction}

\begin{figure}[t]
    \centerline{\includegraphics[clip, width=\columnwidth]{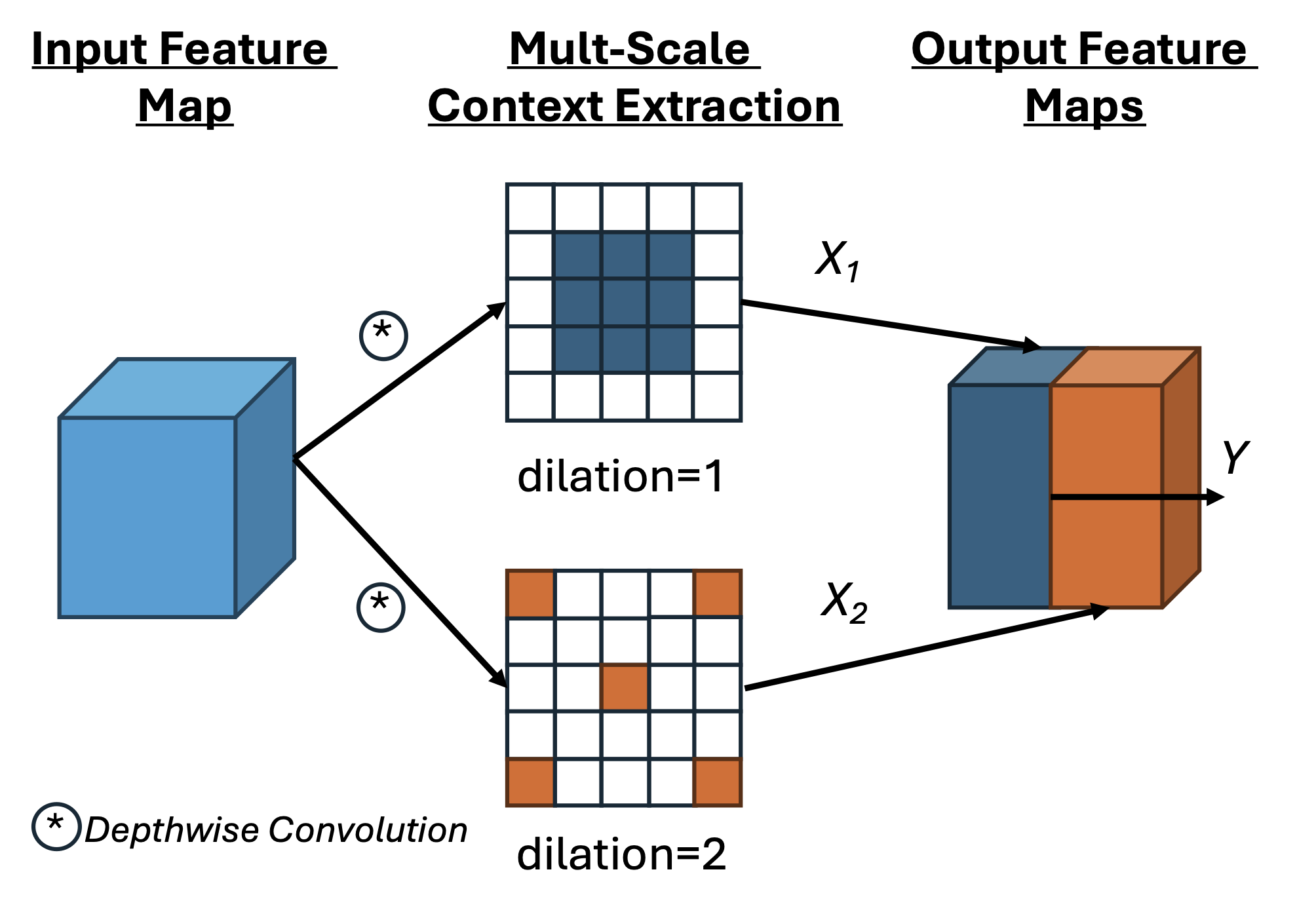}}
    \caption{Multi-Scale Context Extraction through dilated convolution branches.}
    \label{fig:conconv}
\end{figure}

Multi-scale features are extracted by capturing multi-scale context by through parallel branches of dilated convolutions with different rates as shown in Fig. \ref{fig:conconv}. In principle many parallel branches are possible but to reduce computations complexity it is set only to 2. Multi-scale features are extracted to capture both local details and more global context simultaneously. Features are aggregated and dimensionality is increased via simple concatenation which does not incur extra computational cost. Finally, the dilated convolutions are implemented in a depthwise fashion to reduce computational demand leading to independent feature maps operating at effectively different scales. The core operations are as follows:

\begin{align}
& \tensor{X}_{1} = DilatedDepthConv_{d=1}(\tensor{X})\\
& \tensor{X}_{2} = DilatedDepthConv_{d=2}(\tensor{X})\\
& \tensor{Y} = concat([\tensor{X}_{1},\tensor{X}_{2}])
\end{align}

\noindent where $DilatedDepthConv(\cdot)_{d}$ (DDC for short) refers to the dilated convolution operations with dilation rate of $d=1$ (i.e., a canonical $3\times3$ kernel), and with dilation rate of $d=2$ (i.e., effective $5\times5$ receptive field with $9$ parameters). $concat[\cdot,\cdot]$ refers to the channel-wise feature aggregation. In this case concatenating features from different dilated convolutions allows diverse representation of the input that can improve discrimination ability. This operation is followed by a normalization operation and activation which in this case are batch normalization and LeakyRelu respectively.

\subsection{Dilation Shuffling}

\begin{figure}[t]
    \centerline{\includegraphics[clip, width=\columnwidth]{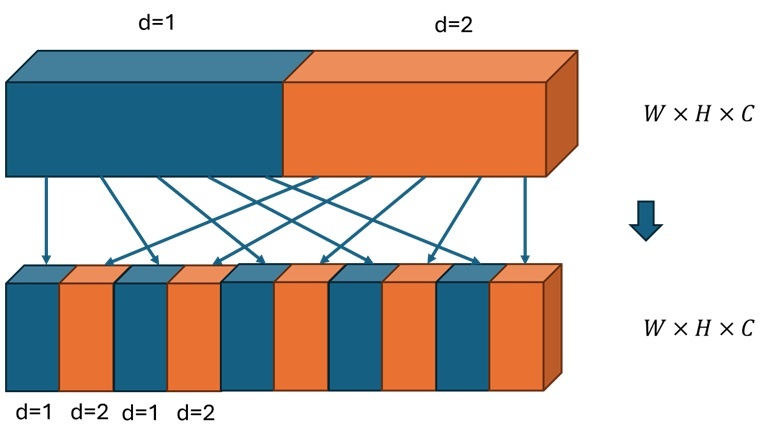}}
    \caption{Channel shuffle to mix channels of different dilation-rate convolutions.}
    \label{fig:shuffle}
\end{figure}

The channel shuffle operation is used to interleave channels originating from possibly multiple feature maps produced by convolution branches (2 in our case), with different dilation rates (as shown in Fig. \ref{fig:shuffle}). After concatenation along the channel dimension, the channels are reshaped into groups corresponding to the different dilation branches and then permuted so that channels $C_i$ from distinct dilation rates are evenly mixed. This reordering breaks the isolation between dilation-specific feature groups and promotes cross-dilation information exchange from different effective scales which is necessary for operating in dynamic and varying environments. This module does not add extra learnable parameters or significant computational overhead.

\begin{figure}[t]
    \centerline{\includegraphics[clip, width=\columnwidth]{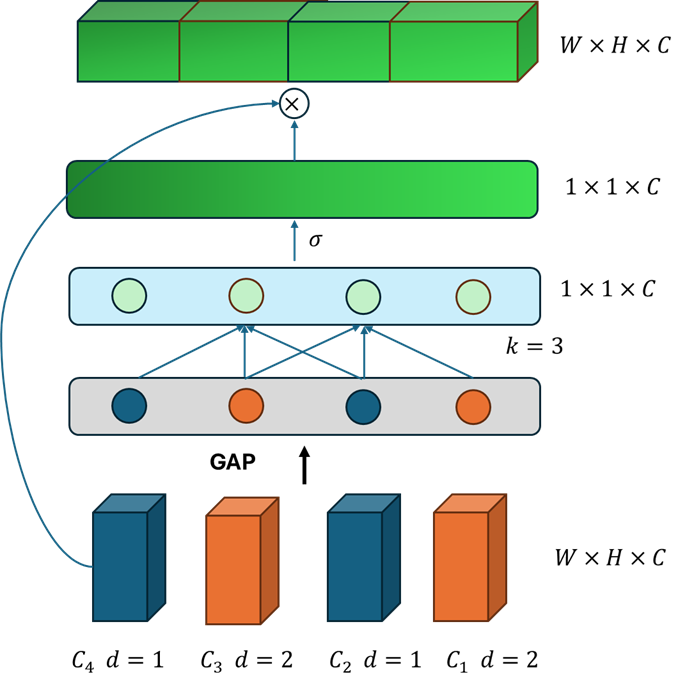}}
    \caption{Efficient Channel Attention to fuse Multi-Scale Context of dilated convolution branches.}
    \label{fig:eca}
\end{figure}

\subsection{Efficient Channel Attention for Cross-Dilation Mixing}

Following channel shuffling, Efficient Channel Attention (ECA) \cite{wang2020eca} is applied to facilitate adaptive cross-dilation mixing as shown in Fig. \ref{fig:eca}. The main idea is to adapt the emphasis of specific channels based on their neighbors stemming from different dilation rates which look at different effective scales. Instead of relying on fully connected layers, ECA employs a lightweight one-dimensional convolution on the globally pooled channel vector to capture local inter-channel dependencies. In the context of multi-rate dilated features, this mechanism allows the network to selectively emphasize or suppress channels originating from different dilation branches based on their contextual relevance. As a result, informative features at appropriate receptive field scales are dynamically weighted, strengthening cross-scale interactions while preserving efficiency by avoiding additional parameters and heavy computation. The introduction of ECA-based dilation channel mixing contributes to a $\sim1\%$ improvement enabling to surpass/match state of the art accuracy as it will be shown in Sec. \ref{sec:deployment}.

\subsection{Overall Architecture}

Based on the main operations we instantiate a small \netname architecture by following a typical conventional architecture of utilizing a stem block, followed by a series of stacked main blocks. Batch normalization is utilized throughout, and LeakyRelu is used as the activation function. Detailed description is shown in Table \ref{tab:architecture}.

\begin{table}[t]
\caption{Stage-wise feature map evolution of the proposed model (input: $240\times240$).}
\centering
\resizebox{\columnwidth}{!}{%
\begin{tabular}{c|c|c}
\hline
\textbf{Stage} & \textbf{Output Size (C$\times$H$\times$W)} & \textbf{Operations} \\
\hline\hline
Input & $3 \times 240 \times 240$ & -- \\ \hline
Stem & $120 \times 120$ & Strided Conv 	$\rightarrow$ BN$\rightarrow$ LeakyReLU \\ \hline
Block 1 & $6 \times 120 \times 120$ & $2\times$DDC-Conv$\rightarrow$ Concat$\rightarrow$ Shuffle$\rightarrow$ ECA Fusion \\ \hline
Block 2 & $8 \times 60 \times 60$ & $2\times$Strided DDC-Conv$\rightarrow$ Concat$\rightarrow$Shuffle$\rightarrow$  Fusion \\ \hline
Block 3 & $14 \times 60 \times 60$ & $2\times$DDC-Conv$\rightarrow$ Concat$\rightarrow$Shuffle$\rightarrow$  ECA Fusion\\ \hline
Block 4 & $38 \times 30 \times 30$ & $2\times$Strided DDC-Conv$\rightarrow$ Concat$\rightarrow$Shuffle$\rightarrow$  ECA Fusion \\ \hline
Block 5 & $56 \times 15 \times 15$ & $2\times$Strided DDC-Conv$\rightarrow$ Concat$\rightarrow$Shuffle$\rightarrow$  ECA Fusion \\ \hline
Block 6 & $56 \times 15 \times 15$ & $2\times$DDC-Conv$\rightarrow$ Concat$\rightarrow$Shuffle$\rightarrow$  ECA Fusion \\ \hline
Block 7 & $62 \times 15 \times 15$ & $2\times$DDC-Conv$\rightarrow$ Concat$\rightarrow$Shuffle$\rightarrow$  ECA Fusion \\ \hline
Head & $N_c \times 1 \times 1$ & $1{\times}1$ Conv$\rightarrow$ GAP \\ \hline
\end{tabular}
}
\label{tab:architecture}
\end{table}

\section{Experiments and Results}

Various experiments are performed to validate the models performance in terms of accuracy against relevant works found in the literature focusing on smaller tiny and mobile style models. Furthermore, smaller models are compared in real deployment scenarios using an NVIDIA platform suitable for edge AI, robotics, and autonomous machine applications. 

\subsection{Training Regime}

Similar to other works \cite{rossi2025takunet,shianios2024direcnetv2} the proposed model is trained for 300 epochs. An initial learning rate of $1e-2$ is used and goes down to $1e-5$. Cosine decay is applied to gradually reduce the learning rate. AdamW optimizer is selected with default parameters. Standard augmentation practices are also applied like random resizing, brightness, hue, saturation adjustments and minor scaling, and rotation. Further, a batch-size of 256 is used and weight decay of $1e-5$. After 300 epochs the best model weights are selected that produced the higher validation value, effectively implementing an early-stopping criterion. The experiments were carried out on a Linux system using 4 RTX 5000 Ada 24GB Graphics Processing Units, with 200GB system RAM and CUDA version 12.9. PyTorch 2.7.0 is used as the deep learning framework.

\subsection{Dataset}

\begin{figure}[t]
    \centerline{\includegraphics[clip, width=\columnwidth]{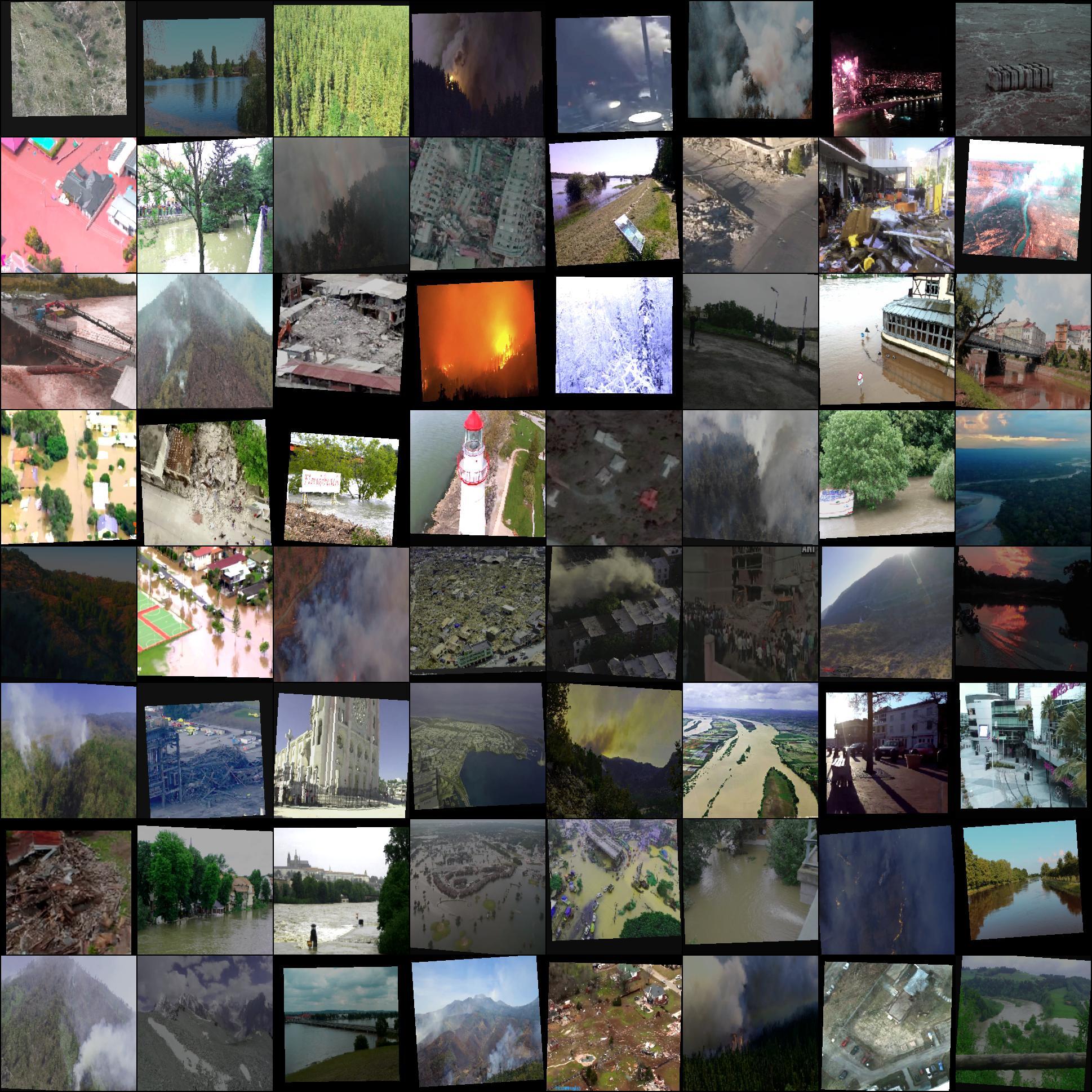}}
    \caption{Sample of images in the AIDERV2 database depicting various images and augmentations applied to the disaster images.}
    \label{fig:aiderv2}
\end{figure}

Two datasets are used for comparisons with related works. AIDERv2 (Aerial Image Dataset for Emergency Response Applications) \cite{shianios2024direcnetv2} is an aerial imagery dataset designed for deep-learning-based disaster detection. It contains 16,723 images categorized into four classes, earthquake/collapsed buildings, floods, wildfires, and normal scenes, with predefined train, validation, and test splits. The dataset extends the original AIDER dataset \cite{kyrkou2020emergencynet} with additional images sourced from other open datasets and frames extracted from YouTube videos. AIDERv2 is intended as a benchmark resource for developing and evaluating models for automated natural disaster recognition in aerial imagery. Example images are shown in Fig. \ref{fig:aiderv2} while statistics are shown in Table \ref{tab:data_desc}. Further the smaller yet relevant AIDER \cite{kyrkou2020emergencynet} dataset is also used, which contains images of emergency situations such as floods, collapsed buildings, traffic accidents, and fires, as a additional source of comparison. 

\begin{table}[!h]
\caption{Proportion of images in each class within the train, validation, and test set.}

\centering
\scalebox{0.95}{%
\begin{tabular}[c]{ c | c | c | c | c | c}
\hline
\centering
& \textbf{Earthquakes} & \textbf{Floods} & \textbf{Wildfire/Fire} & \textbf{Normal} & \textbf{Total} \\[1ex]
\hline
\textbf{Train} & 1927 & 4063 & 3509 & 3900 & 13399 \\
\textbf{Validation} & 239 & 505 & 439 & 487 & 1670 \\
\textbf{Test} & 239 & 502 & 436 & 477 & 1654 \\
\hline
\textbf{Total} & 2405 & 5070 & 4384 & 4864 & \textbf{16723} \\
\hline
\end{tabular}
}
\label{tab:data_desc}
\end{table}

\subsection{Evaluation metrics}

\subsubsection{F1 Score}

Similarly to [24, 33], to validate the accuracy of the method proposed herein, we use the F1-score metric.

\begin{align}
F_1 &= 2 \cdot \frac{P \cdot R}{P + R} \\[1mm]
P &= \frac{\text{TP}}{\text{TP} + \text{FP}} \quad \text{(Precision)} \\[1mm]
R &= \frac{\text{TP}}{\text{TP} + \text{FN}} \quad \text{(Recall)}
\end{align}

\subsubsection{Frames-per-second} Overall latency is compute through frames-per-second (FPS) as follows:

\begin{align}
\mathrm{FPS}_{N_I} = \frac{N_I}{t_{N_I} - t_{1}}
\end{align}

\noindent where $N_I$ is the number of images measured. In our case we set it to 1000.

\begin{table*}[t]
	
	\caption{Comparison of small models in terms of Parameters, FLOPS, Model Size, and F1 Accuracy on the AIDERv2 dataset.}
	\begin{center}
		\resizebox{0.7\textwidth}{!}{%
			\begin{tabular}{c|c|c|c|c} 
				\hline
				\textbf{Model} & \textbf{Parameters} & \textbf{FLOPS} & \textbf{Model Size} & \textbf{F1} \\
				& \textbf{($\times10^3$)} ↓ & \textbf{($\times10^6$)} ↓  & \textbf{(MB)} ↓ & \textbf{(\%)} ↑\\
				\hline \hline
				\textit{MobileNetV2} \cite{sandler2018mobilenetv2} & 2230  & 330 & 8.92  & 89.3 \\ \hline
				\textit{MobileViT xs} \cite{mehta2021mobilevit}& 1930 & 710 & 7.74  & 83.7 \\ \hline
				\textit{ShuffleNet V2} \cite{ma2018shufflenet} & 1260 & 150 & 5.03 & 85.2 \\ \hline
				\textit{MobileViT V2 0.50} \cite{mehta2206separable} & 1110 & 360 & 4.46 & 83.4 \\ \hline
				\textit{MobileViT xxs} \cite{mehta2021mobilevit} & 950 & 260 & 3.81 & 85.9 \\ \hline
				\textit{MobileNet V3 Small} \cite{howard2019searching} & 930 & 60 & 3.72  & 86.8 \\ \hline
				\textit{DirecNet V2} \cite{shianios2024direcnetv2} & 800 & 1090 & 3.20 & \textbf{96.4}\\ \hline
				\textit{SqueezeNet} \cite{iandola2016squeezenet} & 730 & 260 & 2.90 & 84.5 \\ \hline
				\textit{EmergencyNet} \cite{kyrkou2020emergencynet} & 90 & 61 & 0.36  & 95.2\\ \hline
				\textit{TinyEmergencyNet} \cite{mogaka2024tinyemergencynet} & 40 & 31 & 0.16 & 92.8 \\ \hline
				\textit{TakuNet} \cite{rossi2025takunet} & 37  & 31 & 0.15  & 95.8 \\ \hline
				\textit{\textbf{\netnamenospace} (proposed)} & \textbf{23} & \textbf{7} & \textbf{0.1} & 96.1 \\
				\hline
			\end{tabular}
		}
		
		\label{tab:models_comp}
	\end{center}
\end{table*}

\begin{table}[t]
	
	\caption{Comparison of small models in terms of F1 Accuracy on the AIDER dataset.}
	\begin{center}
		\resizebox{\columnwidth}{!}{%
			\begin{tabular}{c|c|c|c|c} 
				\hline
				\textbf{Model} & \textbf{Parameters} & \textbf{FLOPS} & \textbf{Model Size} & \textbf{F1} \\
				& \textbf{($\times10^3$)} ↓ & \textbf{($\times10^6$)} ↓  & \textbf{(MB)} ↓ & \textbf{(\%)} ↑\\
				\hline \hline
				\textit{EmergencyNet} \cite{kyrkou2020emergencynet} & 90 & 77 & 0.36  & 93.6\\ \hline
				\textit{TinyEmergencyNet} \cite{mogaka2024tinyemergencynet} & 40 & 36 & 0.16 & 89.5 \\ \hline
				\textit{TakuNet} \cite{rossi2025takunet} & 37  & 35 & 0.15  & 94.3 \\ \hline
				\textit{\textbf{\netnamenospace} (proposed)} & \textbf{23} & \textbf{7} & \textbf{0.1} & \textbf{95.0} \\
				\hline
			\end{tabular}
		}
		
		\label{tab:models_comp2}
	\end{center}
\end{table}

\subsection{Experimental Results}

\subsubsection{Comparisons}
Table \ref{tab:models_comp} shows results as reported in prior works. Overall, it highlights a clear efficiency-accuracy trade-off across small models, with several points worth mentioning. DirecNetV2, despite having only 800k parameters, incurs the highest FLOPs (1090M). Its strong accuracy (96.4\%) is achieved through computationally intensive operations such as the self-attention which is also memory demanding, constraining its use in energy-constrained deployments. EmergencyNet, TakuNet, and the proposed \netname all achieve $\geq$95\% accuracy with two orders of magnitude fewer parameters than mainstream mobile backbones. This strongly indicates that architectural specialization for emergency scene recognition is more impactful than scaling down general-purpose CNNs or ViTs. \netname reaches 96.1\% accuracy with only 23k parameters, 7M FLOPs, and a 0.1 MB for weight storage. While it is marginally below DirecNetV2 in absolute accuracy (-0.3\%), it is $\sim150\times$ smaller in FLOPs and $\sim35\times$ smaller in parameters, placing it clearly ahead on any resource-constrained model seting a new state of the art in efficiency with minimal accuracy drop.

Furthermore the results are validated using the AIDER dataset which while smaller in size contains an additional class. Results are shown in Table \ref{tab:models_comp2}. Compared to results on AIDERv2, all models experience some degradation, which is expected given the reduced data amount and increased class cardinality. However, the magnitude of the drop differs substantially. TinyEmergencyNet suffers the largest decline (to 89.5\%), suggesting limited robustness when class boundaries become more complex under data scarcity. EmergencyNet (90k params) does not clearly outperform much smaller architectures indicating that simply increasing capacity might be insufficient. TakuNet and the proposed \netnamenospace, both under 40k parameters, outperform EmergencyNet despite being substantially smaller. \netname achieves the highest accuracy (95.0\%) while also being the smallest and cheapest model by a wide margin (7M FLOPs). Overall, the results strengthen the claim that \netname is not only efficient but also more sample-efficient and resilient than existing lightweight baselines.

\subsubsection{Deployment Aspects}\label{sec:deployment}

Additionally, smaller models namely the proposed one and TakuNet, are evaluated  in terms of frames per second (FPS), by deploying them on NVIDIA Jetson Orin AGX device using both the GPU and CPU available on the platform. The FPS are calculated as the mean latency over multiple runs as shown in Eq. 7, with batch size equal to 1 to simulate streaming camera feed, and image size of $240\times240$ pixels, without applying any optimizations such as ONNX, TensorRT conversion, pruning, or quantization. This permits us to assess the effectiveness of the architectural decisions before relying on hardware optimizations for additional gains. 

The proposed model achieves 258 FPS on the GPU, substantially outperforming TakuNet at 145 FPS. This $\sim1.8\times$ speedup is consistent with the gap in FLOPs (7M vs. $\sim$35M) and confirms that the architectural savings translate almost linearly into real-time throughput on embedded GPUs.

On the ARM-based CPU, the proposed model reaches 46 FPS, vastly outperforming TakuNet at 7.6 FPS ($\sim6\times$ speedup). This result is significant, as it further emphasizes that the architectural choices of dilated depth-wise convolutions, early downsampling, and concatenation-based channel expansion are suitable for CPU inference which is computationally more limited than a GPU. Similarly to the GPU no additional optimizations were performed further indicating that the performance is stemming from the architectural choices. 

The proposed model also demonstrates 4.4 MB peak activation memory, compared to 32.48 MB for TakuNet. This $\sim7\times$ reduction is significant as it enables deployment on platforms with limited memory but also is well-suited to run alongside other perception modules as it reduces memory demands and is also capable of multi-stream inference.

\section{Conclusion}

This work presented \netnamenospace, an ultra-compact neural network designed for efficient aerial image disaster scene recognition under strict resource constraints. Across two datasets, \netname consistently achieved near or higher state-of-the-art accuracy while using orders of magnitude fewer parameters, FLOPs, and memory than existing lightweight models. Notably, it matched or surpassed the performance of significantly larger architectures that utilize enhanced features such as self-attention. Beyond accuracy metrics, extensive deployment experiments on the NVIDIA Jetson Orin AGX demonstrated that the proposed model’s efficiency translates directly into practical deployment. \netname delivered the highest inference throughput on both GPU and CPU, together with a substantially reduced memory footprint, confirming its suitability for real-time, embedded, and internet-denied aerial emergency response systems. Overall, \netname establishes a new reference point for accuracy-efficiency trade-offs for tinyML in emergency recognition.

\section*{Acknowledgements}
This work was supported by the European Union’s Horizon Europe research and innovation programme under grant agreement No. 101168067 (GuardAI). Views and opinions expressed are however those of the author(s) only and do not necessarily reflect those of the European Union. Neither the European Union nor the granting authority can be held responsible for them.

\bibliographystyle{plain}
\bibliography{ref.bib}

\begin{thebibliography}{10}

\bibitem{bouguettaya2019survey}
Abdelmalek Bouguettaya, Ahmed Kechida, and Amine~Mohammed TABERKIT.
\newblock A survey on lightweight cnn-based object detection algorithms for
  platforms with limited computational resources.
\newblock {\em International Journal of Informatics and Applied Mathematics},
  2(2):28--44, 2019.

\bibitem{chen2017deeplab}
Liang-Chieh Chen, George Papandreou, Iasonas Kokkinos, Kevin Murphy, and Alan~L
  Yuille.
\newblock Deeplab: Semantic image segmentation with deep convolutional nets,
  atrous convolution, and fully connected crfs.
\newblock {\em IEEE transactions on pattern analysis and machine intelligence},
  40(4):834--848, 2017.

\bibitem{dosovitskiy2020image}
Alexey Dosovitskiy, Lucas Beyer, Alexander Kolesnikov, Dirk Weissenborn,
  Xiaohua Zhai, Thomas Unterthiner, Mostafa Dehghani, Matthias Minderer, Georg
  Heigold, Sylvain Gelly, et~al.
\newblock An image is worth 16x16 words: Transformers for image recognition at
  scale.
\newblock {\em arXiv preprint arXiv:2010.11929}, 2020.

\bibitem{ghali2021wildfire}
Rafik Ghali, Moulay~A Akhloufi, Marwa Jmal, Wided Souidene~Mseddi, and Rabah
  Attia.
\newblock Wildfire segmentation using deep vision transformers.
\newblock {\em Remote Sensing}, 13(17):3527, 2021.

\bibitem{habash2025aerial_od_survey}
Nadin Habash, Ahmad Abu~Alqumsan, and Tao Zhou.
\newblock Recent real-time aerial object detection approaches, performance,
  optimization, and efficient design trends for onboard performance: A survey.
\newblock {\em Sensors}, 25(24):7563, 2025.

\bibitem{howard2019searching}
Andrew Howard, Mark Sandler, Grace Chu, Liang-Chieh Chen, Bo~Chen, Mingxing
  Tan, Weijun Wang, Yukun Zhu, Ruoming Pang, Vijay Vasudevan, et~al.
\newblock Searching for mobilenetv3.
\newblock In {\em Proceedings of the IEEE/CVF international conference on
  computer vision}, pages 1314--1324, 2019.

\bibitem{iandola2016squeezenet}
Forrest~N Iandola, Song Han, Matthew~W Moskewicz, Khalid Ashraf, William~J
  Dally, and Kurt Keutzer.
\newblock Squeezenet: Alexnet-level accuracy with 50x fewer parameters and< 0.5
  mb model size.
\newblock {\em arXiv preprint arXiv:1602.07360}, 2016.

\bibitem{khudayberdiev2022fire}
Otabek Khudayberdiev, Jiashu Zhang, Ahmed Elkhalil, and Lansana Balde.
\newblock Fire detection approach based on vision transformer.
\newblock In {\em International Conference on Adaptive and Intelligent
  Systems}, pages 41--53. Springer, 2022.

\bibitem{krizhevsky2017imagenet}
Alex Krizhevsky, Ilya Sutskever, and Geoffrey~E Hinton.
\newblock Imagenet classification with deep convolutional neural networks.
\newblock {\em Communications of the ACM}, 60(6):84--90, 2017.

\bibitem{9979171}
Christos Kyrkou, Panayiotis Kolios, Theocharis Theocharides, and Marios
  Polycarpou.
\newblock Machine learning for emergency management: A survey and future
  outlook.
\newblock {\em Proceedings of the IEEE}, 111(1):19--41, 2023.

\bibitem{9025436}
Christos Kyrkou and Theocharis Theocharides.
\newblock Deep-learning-based aerial image classification for emergency
  response applications using unmanned aerial vehicles.
\newblock In {\em 2019 IEEE/CVF Conference on Computer Vision and Pattern
  Recognition Workshops (CVPRW)}, pages 517--525, 2019.

\bibitem{kyrkou2020emergencynet}
Christos Kyrkou and Theocharis Theocharides.
\newblock Emergencynet: Efficient aerial image classification for drone-based
  emergency monitoring using atrous convolutional feature fusion.
\newblock {\em IEEE Journal of Selected Topics in Applied Earth Observations
  and Remote Sensing}, 13:1687--1699, 2020.

\bibitem{lee2023watt}
Gao~Yu Lee, Tanmoy Dam, Md~Meftahul Ferdaus, Daniel~Puiu Poenar, and Vu~N
  Duong.
\newblock Watt-effnet: A lightweight and accurate model for classifying aerial
  disaster images.
\newblock {\em IEEE Geoscience and Remote Sensing Letters}, 2023.

\bibitem{ma2018shufflenet}
Ningning Ma, Xiangyu Zhang, Hai-Tao Zheng, and Jian Sun.
\newblock Shufflenet v2: Practical guidelines for efficient cnn architecture
  design.
\newblock In {\em Proceedings of the European conference on computer vision
  (ECCV)}, pages 116--131, 2018.

\bibitem{mehta2206separable}
S~Mehta and M~Rastegari.
\newblock Separable self-attention for mobile vision transformers. arxiv 2022.
\newblock {\em arXiv preprint arXiv:2206.02680}.

\bibitem{mehta2021mobilevit}
Sachin Mehta and Mohammad Rastegari.
\newblock Mobilevit: light-weight, general-purpose, and mobile-friendly vision
  transformer.
\newblock {\em arXiv preprint arXiv:2110.02178}, 2021.

\bibitem{mogaka2024tinyemergencynet}
Obed~M Mogaka, Rami Zewail, Koji Inoue, and Mohammed~S Sayed.
\newblock Tinyemergencynet: a hardware-friendly ultra-lightweight deep learning
  model for aerial scene image classification.
\newblock {\em Journal of Real-Time Image Processing}, 21(2):51, 2024.

\bibitem{munsif2022lightweight}
Muhammad Munsif, Hina Afridi, Mohib Ullah, Sultan~Daud Khan, Faouzi~Alaya
  Cheikh, and Muhammad Sajjad.
\newblock A lightweight convolution neural network for automatic disasters
  recognition.
\newblock In {\em 2022 10th European Workshop on Visual Information Processing
  (EUVIP)}, pages 1--6. IEEE, 2022.

\bibitem{rossi2025takunet}
Daniel Rossi, Guido Borghi, and Roberto Vezzani.
\newblock Takunet: an energy-efficient cnn for real-time inference on embedded
  uav systems in emergency response scenarios.
\newblock In {\em Proceedings of the Winter Conference on Applications of
  Computer Vision}, pages 376--385, 2025.

\bibitem{sandler2018mobilenetv2}
Mark Sandler, Andrew Howard, Menglong Zhu, Andrey Zhmoginov, and Liang-Chieh
  Chen.
\newblock Mobilenetv2: Inverted residuals and linear bottlenecks.
\newblock In {\em Proceedings of the IEEE conference on computer vision and
  pattern recognition}, pages 4510--4520, 2018.

\bibitem{shianios2024direcnetv2}
Demetris Shianios, Panayiotis~S Kolios, and Christos Kyrkou.
\newblock Direcnetv2: A transformer-enhanced network for aerial disaster
  recognition.
\newblock {\em SN Computer Science}, 5(6):770, 2024.

\bibitem{szegedy2015going}
Christian Szegedy, Wei Liu, Yangqing Jia, Pierre Sermanet, Scott Reed, Dragomir
  Anguelov, Dumitru Erhan, Vincent Vanhoucke, and Andrew Rabinovich.
\newblock Going deeper with convolutions.
\newblock In {\em Proceedings of the IEEE conference on computer vision and
  pattern recognition}, pages 1--9, 2015.

\bibitem{wang2020eca}
Qilong Wang, Banggu Wu, Pengfei Zhu, Peihua Li, Wangmeng Zuo, and Qinghua Hu.
\newblock Eca-net: Efficient channel attention for deep convolutional neural
  networks.
\newblock In {\em Proceedings of the IEEE/CVF conference on computer vision and
  pattern recognition}, pages 11534--11542, 2020.

\bibitem{yang2022efficient}
Nathaniel Tan~Sze Yang, Mau-Luen Tham, Sing~Yee Chua, Ying~Loong Lee, Yasunori
  Owada, and Suvit Poomrittigul.
\newblock Efficient device-edge inference for disaster classification.
\newblock In {\em 2022 Thirteenth International Conference on Ubiquitous and
  Future Networks (ICUFN)}, pages 314--319. IEEE, 2022.

\end{thebibliography}

\end{document}